\documentclass[letterpaper, 10 pt, conference]{ieeeconf}  

\IEEEoverridecommandlockouts 
\usepackage{graphicx}
\usepackage{comment}
\usepackage{amsmath,amssymb} 
\usepackage[T1]{fontenc}
\usepackage{aecompl}
\usepackage{color}
\usepackage{times}
\usepackage{epsfig}
\usepackage{float}
\usepackage{booktabs}
\usepackage{multirow}
\usepackage{caption}
\usepackage{subfigure}
\usepackage{adjustbox}
\usepackage{xspace}
\usepackage{dsfont}
\usepackage{bbm}
\usepackage{framed}
\usepackage{color}
\usepackage{url}
\definecolor{shadecolor}{rgb}{0.1, 0.1, 0.9}
\newcommand{\model}{IPC\xspace}

\newcommand\minisection[1]{\vspace{1mm}\noindent \textbf{#1}}

\newcommand{\vect}{\bold}   
\newcommand{\spac}{\mathcal}    
\newcommand{\subsct}{\text}  

\newcommand{\func}{\mathit}    
\newcommand{\action}{\vect{a}}  
\newcommand{\state}{\vect{s}}   
\newcommand{\observ}{\vect{o}}   
\newcommand{\actcomb}{\vect{A}} 
\newcommand{\cfunc}{\func{C}}

\author{Jinkun Cao$^{1}$,  Xin Wang$^{2}$,  Trevor Darrell$^{2}$ and Fisher Yu$^{3}$ \\ $^{1}$ Carnegie Mellon University  $^{2}$ University of California, Berkeley $^{3}$ ETH Zurich\\ {\tt\small jinkunc@andrew.cmu.edu}, {\tt\small \{xinw, trevor\}@eecs.berkeley.edu}, {\tt\small i@yf.io}}

\begin{document}

\title{Instance-Aware Predictive Navigation in Multi-Agent Environments}

\maketitle

\IEEEpeerreviewmaketitle

\begin{abstract}
In this work, we aim to achieve efﬁcient end-to-end learning of driving policies in dynamic multi-agent environments. Predicting and anticipating future events at the object level are critical for making informed driving decisions. We propose an Instance-Aware Predictive Control (IPC) approach, which forecasts interactions between agents as well as future scene structures. We adopt a novel multi-instance event prediction module to estimate the possible interaction among agents in the ego-centric view, conditioned on the selected action sequence of the ego-vehicle. To decide the action at each step, we seek the action sequence that can lead to safe future states based on the prediction module outputs by repeatedly sampling likely action sequences. We design a sequential action sampling strategy to better leverage predicted states on both scene-level and instance-level. Our method establishes a new state of the art in the challenging CARLA multi-agent driving simulation environments without expert demonstration, giving better explainability and sample efficiency. 

\end{abstract}

\section{Introduction}

Over the past years, machine vision and decision-making systems have approached or surpassed human-level
performance in vision and robotics
applications due to the emergence of deep
learning methods~\cite{krizhevsky2012imagenet}. However, learning drive autonomously remains one of the most desirable but notoriously challenging problems, with high requirements on reliability, explainability, and data efficiency~\cite{pomerleau1989alvinn,spc,huval2015empirical}.

One approach to learning a driving policy is imitation learning~\cite{CIRL,experdriving1,expertdriving2,expertdriving3}, where the policy is fully supervised by expert demonstrations. 
However, it may
be unreliable when facing scenarios outside demonstrations as expert demonstrations provide only
examples of safe driving instead of recovering from mistakes. Besides, demonstrations are hard to collect, even in some simulation environments. On the other hand, reinforcement learning-based agents can explore the world and learn from mistakes without expert demonstrations~\cite{modelbased1,modelbased2}. But they suffer severely from sample
inefficiency as they rely on sparse reward signals from the environment and require orders of magnitude more data than humans~\cite{silver2017mastering}. It brings the utmost challenges to drive in a multi-agent environment, where the scene changes are more complex and even adversarial to the ego-vehicle.

We aim to have reliable driving policies by exploring multi-agent
environments, combining the deep representation of imitation with the
reinforcement from exploratory experiences. To achieve reliability, we need to predict the consequences of the selected actions. At the same time, we want to learn from experiences to improve data efficiency. Learning from good experience allows self-imitation and eliminates the necessity of expert demonstrations.

\begin{figure}
    \centering
    \includegraphics[width=\linewidth]{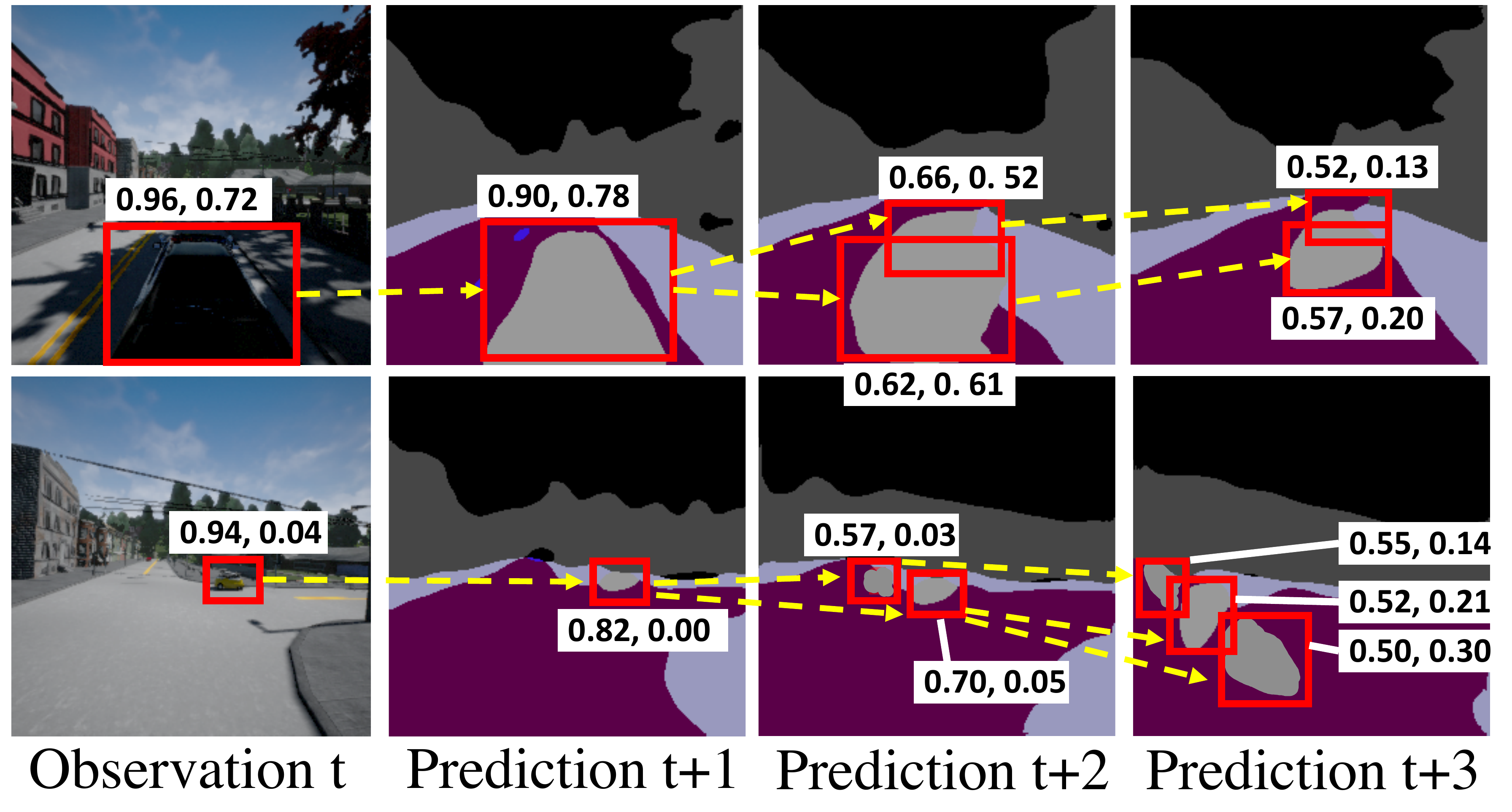}
    \caption{Multi-instance event prediction in ego-centric view. It predicts both possible instance locations and the probability of instance-level events. The predicted instance locations on a frame sequence build a coarse estimation of possible trajectories. For each possible location, the model outputs confidence of localization and chance of collision, $(c, \phi)$. 
    \vspace{-8mm}}
    \label{MILP}
\end{figure}

We propose Instance-Aware Predictive Control (\model), a decision-making method with future event prediction at both scene and instance levels. \model~forecasts the motions of the other agents in the environment as well as the scene structure changes. A key component is a \textit{multi-instance event prediction} module. It estimates the possible locations of the other agents in the dynamic environments within a finite horizon, conditioned on the past and present observations and the selected action sequence of the controlled ego-agent. The likely locations of the other agents indicate possible future events, such as collisions. Our agent makes action decisions by considering sampled actions' consequences. The predicted visual structure helps to explain the agent's decision.

To determine the action, \model~seeks an optimal action sequence within a finite horizon by repeatedly sampling possible action sequences. The prediction module can tell which sequence leads to safe future states, and the agent executes the action for the first step from the sequence. We design a
\textit{sequential action selection} strategy, selecting actions to optimize the chances of safe scene-level and instance-level events. It improves the efficiency of planning with noisy prediction. Our experiments show that the strategy is helpful to select action sequences more efficiently.

\model establishes a new state of the art in the challenging CARLA simulator and the more realistic GTA V environment.
\model~ improves the sample efficiency and policy
performance without any expert demonstration. Moreover, the instance-level visual prediction enhances the explainability of action decisions. Our method predicts future safety-related events (e.g., collision, offroad, offlane) with 95\%+ accuracy. The codebase is released at \url{https://github.com/SysCV/spc2}. 
\section{Related Work}
Our work connects to the rich literature in reinforcement
learning and imitation learning. In this section, we focus on related works within the scope of autonomous driving.

\minisection{Forecasting and control.}
The state forecasting is usually handled as a Markov Decision Process (MDP), and state forecasting conditional to certain actions could be used in control tasks. Though it is still hard to learn policy directly from pixels, deep models have brought powerful representations for visual tasks. Recurrent networks~\cite{rochan2018future, chiu2020segmenting, sociallstm,bilinearlstmtracking} have shown the power to predict future visual states based on raw visual inputs. However, in multi-agent environments, the uncertainty from other agents makes it harder to infer the underlying distribution of state transition. Explicit multi-hypothesis forecasting~\cite{PRECOG} focuses on this problem but requires full instance trajectories. Instead, we use cues of multi-modal future state possibility from only ego-centric monocular observations. Our method refers to Model Predictive Control (MPC)~\cite{camacho2013model} to control the ego-vehicle by comparing the forecast future states conditional to some selected actions. Pan et al.~\cite{spc} also uses the way to learn driving policy but only in single-agent environments. We are the first to use instance awareness to achieve an efficient driving policy in a complicated multi-agent environment with an explainable decision. 

\minisection{Driving by imitation.}
Imitation learning focuses on imitating policy pattern from expert demonstrations. Previous methods~\cite{imitation1,imitation2, CIRL, experdriving1, expertdriving2} have shown efforts in autonomous driving tasks. However, the demonstrations focus on only correct behaviors, providing no chance for the agent to learn from mistakes. It can lead to dangerous situations as the agent would not know how to recover from accidents. Besides, demonstrations are usually very hard and expensive to collect in the real world, and manually designed demonstrations in simulators usually suffer from a lack of diversity. So imitation learning is also studied together with exploring in experience~\cite{expertdriving3} to relieve its imperfectness. Our method also uses imitation to boost performance, but it uses imperfect pseudo-demonstration from previous ``good experience'', which is called self-imitation.

\minisection{Policy learning from experiences.}
One way to eliminate the necessity of demonstrations is to learn policy from the agent's experiences. Reinforcement learning typically follows this philosophy. Recently, model-based reinforcement learning has been gaining popularity, leveraging deep networks to learn a state transition function~\cite{modelbased1,modelbased2} for action
decisions. Though some methods~\cite{RLdriving1, RLdriving3} has tried to learn policies directly from observations without demonstration, they usually work well only in over-simplified simulation environments~\cite{mnih2013playing, kempka2016vizdoom}. For driving tasks, the existing reinforcement learning method in the realistic simulator, e.g., CARLA~\cite{Carla}, still requires imitating demonstrations~\cite{CIRL}. Besides, compared with visual prediction, deep models still suffer from severe low efficiency when planning action directly from visual signals. 
Aware of these flaws, our method does not directly output action decisions from visual input but by comparing the consequences of sampled action candidates, which better leverages the capacity of deep models.

\section{Instance-Aware Predictive Control}
In this section, we start with an introduction to the problem setup (\ref{sec:setup}) and an overview of the proposed Instance-Aware Predictive Control (IPC) framework (\ref{sec:overview}). We then dive into its two core modules:
Multi-instance Event Prediction (MEP) for forecasting (\ref{sec:mlp}) and Sequential Action Sampling (SAS) for action planning (\ref{sec:sas}).

\begin{figure*}[t]
    \centering
    \includegraphics[width=.85\linewidth]{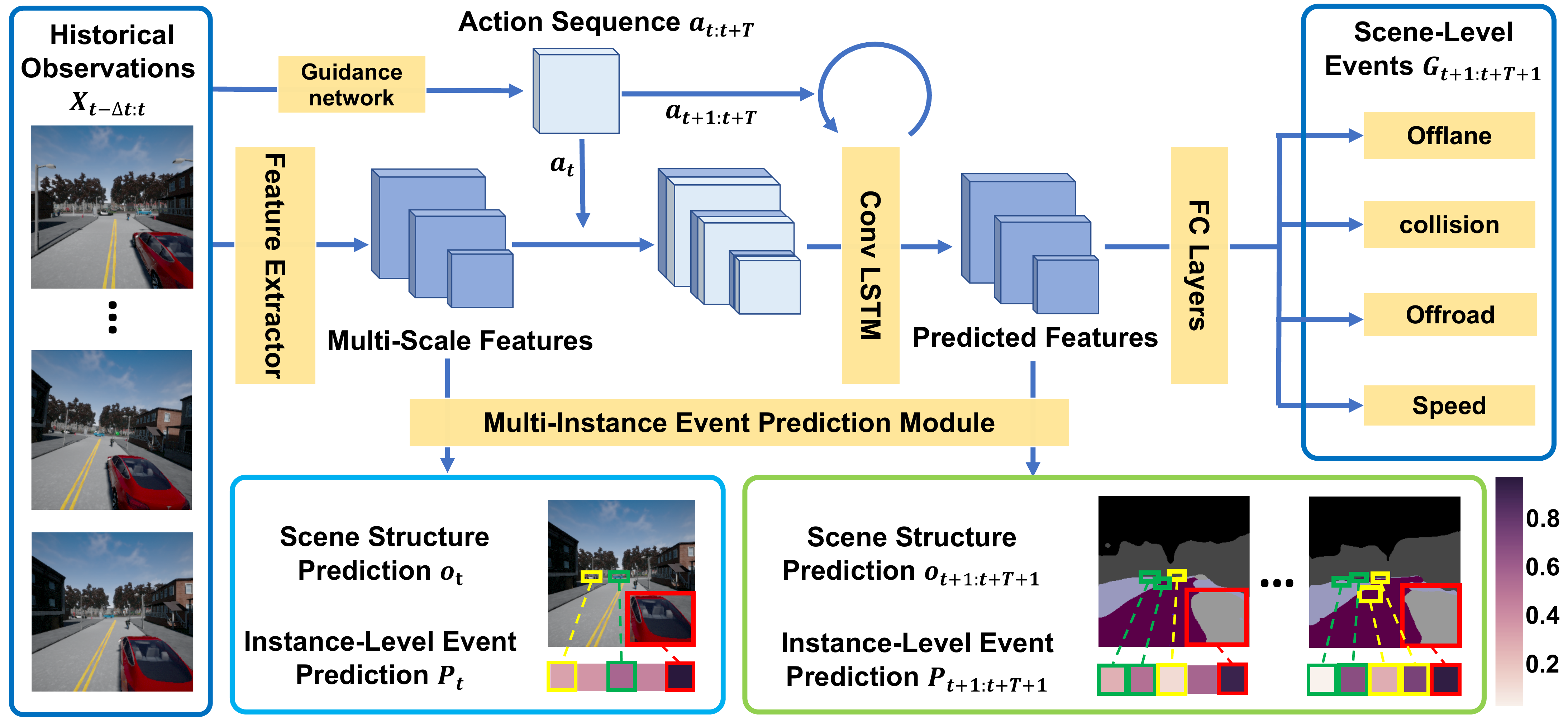}
    \caption{Instance-aware predictive control (\model) framework. Given historical observations, a guidance network helps to sample action sequences in action space. The model predicts both future visual structure and the chance of some events. $\observ$ is the visual observation containing semantic segmentation and instance locations. $G$ is scene-level events. $P$ are instance-level events on each predicted possible instance location. Event prediction brings reference to action selection. Visual structure prediction brings explanation to the action decision. The bottom-right colorbar indicates the probability of instance-level events.}
    \label{fig:framework}
\end{figure*}

\subsection{Problem Setup}
\label{sec:setup}
Following previous works~\cite{spc, MPC1}, we formulate 
the environment transition under Markov Decision Process. With the state space $\spac{S}$ and the action space $\spac{A}$, the transition is $\func{M}: \spac{S} \times \spac{A}^{N} \longrightarrow \spac{S}$ in an N-agent environment,
\begin{equation}
    \state_{t+1} = \func{M}(\state_t, \actcomb_{t}),
    \label{transition}
\end{equation}
where $\actcomb_{t}=(\action^0_{t}, \dots, \action^{N}_{t})$ are the
actions executed by N agents at timestamp $t$, $\action^0_t$ is the action of the ego-vehicle and $\state$ is the partial observation of the world.

We only optimize the ego-vehicle's driving policy $\pi^0$, taking into account the environment and other agents' behaviors. In contrast, other agents adopt a pre-defined policy from the simulator that is unknown to the ego-vehicle. The reward function $\func{R}$ in the MDP rewards the ego-vehicle to drive fast and safely in the environment. That is, the ego-vehicle needs to drive as fast as possible without crashing into other vehicles or obstacles or getting off the drivable area. 

\subsection{Instance-aware Predictive Control Framework}
\label{sec:overview}
\model is a decision-making method with future event predictions at both scene and instance level. Similar to the classic model predictive control, \model  
includes a dynamic model to predict future states and a sampling-based strategy to select the action with respect to the state prediction.

\minisection{Predictive control.} To determine the ego-vehicle's action at timestamp $t$, we adopt a predictive control approach that samples multiple action sequences spanning on timestamps $t:t+T$ and forecasts future states conditioned on the sampled action sequences. \model inherits the merits of the model predictive control framework that implements the action of the current timestamp $t$ while keeping future timestamps in account. In \model, we train a dynamic model $M^*$ that predicts future states with the observation of the current state and action sequence. Once the dynamic model is learned, we can sample the action sequences conditioned on the future state forecasting and adopt a cost function $C$ to optimize the action sampling. That is, 
\begin{equation}
    \action^{0}_{t:t+T} = \mathop{\arg\min}\limits_{\action^{0}_{t:t+T}}  \sum_{\tau=t}^{t+T} \cfunc (\func{M}^{*}(\state_{\tau}, \actcomb_{\tau})), 
    \label{eq:act}
\end{equation}
where $T$ is the length of horizon for state forecasting.
To better learn the dynamic model, we adopt an episodic learning strategy to train the prediction module where the ground truth only exposes one possibility of the future in one episode. The model observes multiple possible future outcomes caused by different behaviors of other vehicles in the environment by exploring multiple episodes.

\minisection{State forecasting.} A key component in \model is its future events prediction module at both scene and instance level. As shown in Fig.~\ref{fig:framework}, the future prediction module first 
extracts feature representations from a sequence of historical observations, which encode past dynamics of other agents. 
The visual features are then concatenated with sampled actions and fed into a convolutional LSTM~\cite{convlstm} network to predict features on future $T$ timestamps recurrently. Once the features are obtained, 
\model utilizes a multi-instance event prediction (MEP) module to estimate future visual structure $\observ$ with semantic segmentation and possible locations of other vehicles through object detection. MEP has a RetinaNet~\cite{focalloss}-like network to estimate the chances of instance-level events between the ego-vehicle and other vehicles (e.g., collision) denoted as $P$. \model also includes a fully connected network to predict scene-level events denoted as $G$. The overall state in the dynamic model is represented by a triplet 
\vspace{-1mm}
\begin{equation}
    \state = (\observ, G, P),
    \label{eq:state}
\end{equation}
where $\observ$ provides semantic visual information and $G$ and $P$ are safety-related events or ego-vehicle speed. To note that, although the ego-vehicle is unaware of the policies of other vehicles, 

the observation $\observ$ reflects states of other agents,
which influences the action selection of the ego-vehicle.

State forecasting in a multi-agent environment is challenging due to its dynamic nature, where various behaviors from other agents can lead to uncertain future predictions. To address this issue, we adopt a multiple hypotheses forecasting strategy as shown in Fig.~\ref{fig:mep}, where MEP explicitly takes into account multiple possible configurations of the future states and provides uncertainty scores to assist action selection.

\minisection{Action selection.} We develop an action selection method to ensure the ego-vehicle drives within drivable areas and avoids collisions in the environment. According to Eq.~\ref{eq:act} and \ref{eq:state}, the action selection is conditioned on both the scene-level and the instance-level events which bear more uncertainty from non-deterministic
behaviors of other agents. 
To determine the action sequence efficiently, we design a two-stage sequential action sampling (SAS) strategy to select action as illustrated in Fig.~\ref{actionselection}. SAS assesses sampled action candidates in the first stage and eliminates the unlikely action sequences based on the scene-level predictions only. Then it uses noisy instance-level predictions to determine the final action sequence.

\subsection{Multi-Instance Event Prediction}
\label{sec:mlp}

A core design of the state forecasting in \model is the multi-instance event prediction, which explicitly models the state changes in the multi-agent environment. To achieve it, the multi-instance event prediction (MEP) module utilizes visual perception tasks such as semantic segmentation to capture the scene structure changes and object detection to model inter-agent interaction.

Event prediction in a multi-agent environment is challenging. In addition to recognizing the road structure and following the correct lanes, our agent will also avoid collisions with the other agents. Although we are testing our systems mainly in the simulation environment, we assume we can only control our vehicle, and the policies of those agents are unknown. Therefore, we do not know their exact driving trajectories and speed, and their driving policies are indeed programmed with some degree of randomness. The assumption and challenge are also realistic. When we drive in the real world, we do not know exactly what is in the mind of other drivers, riders, and pedestrians.

Our key insight is that certain motion patterns, despite their stochastic nature, will emerge from a large number of behavioral observations of the agents in the environments. For instance, a car will unlikely stop suddenly without the influence of the scene structures or interference of the other agents, and a pedestrian will likely walk straight on the sidewalks. Also, the object locations across time should be continuous. Therefore, our prediction module, MEP, recognizes the existences of the multiple agents and attempts to learn their motion patterns with deep neural networks. In this work, we represent the instances and their motion with 2D bounding boxes.

\begin{figure}[t]
    \centering
    \includegraphics[width=\linewidth]{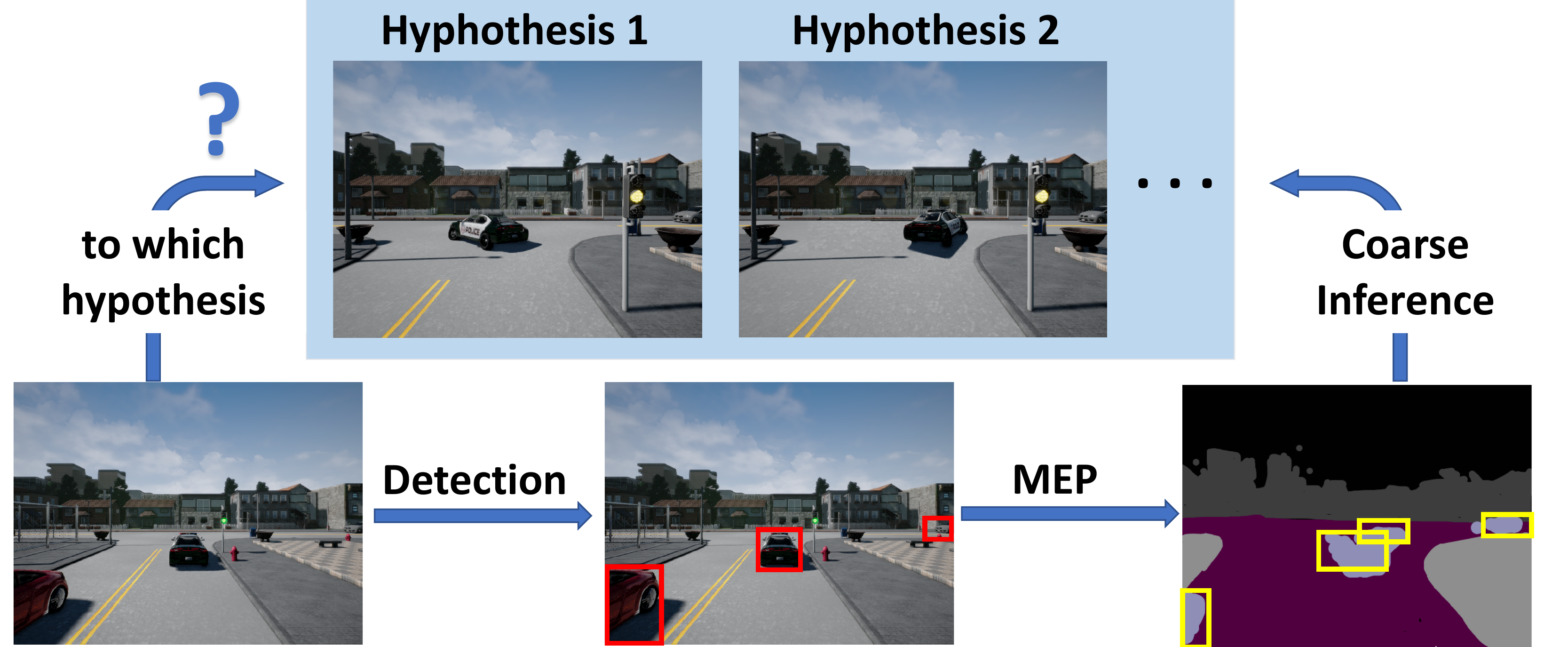}
    \caption{How the prediction of possible instance locations in the Multi-instance Event Prediction (MEP) builds a coarse inference for different hypotheses.
    }
    \label{fig:mep}
\end{figure}

In addition, MEP models the behavioral randomness of the future through multiple hypotheses, similar to using modes to represent a distribution. A car may want to speed up or make a turn, leading to uncertainty in the future states. MEP represents the uncertainty with likely future positions of the other agents shown in Fig.~\ref{fig:mep}. Also, MEP provides the uncertainty scores of those possible future locations. 

Semantic segmentation can provide dense supervision to the driving policy learning in addition to the sparse rewards from the environment, which improves the sample efficiency and the explainability of action decisions, as shown in the prior work~\cite{spc}. However, the original design of SPC only handles static scene structures with one possible future state hypothesis. It also fails to capture the dynamic scene changes in a multi-agent environment, where instance-level events (e.g., collisions between vehicles) dominate the scene changes. Therefore, \model uses MEP to predict extra instance-level events based on future state and scene structure estimation. The dense supervision for semantic segmentation and object detection provides richer training signals and improves the sample efficiency and explainability.

\minisection{Training.} To train MEP, we adopt an episodic training strategy where episodes are sampled from a replay buffer, and the loss is defined over a finite horizon of $T$. That is,  
\vspace{-2mm}
\begin{equation}
    \mathcal{L}_{\subsct{D}} = \sum^T_{t=1} (w_t \mathcal{L}_{\subsct{d}} + \mathcal{L}_{\subsct{g}}),
    \label{location_loss}
\end{equation}      

where $\mathcal{L}_\subsct{d}$ is the focal loss~\cite{focalloss} for location estimation and $\mathcal{L}_\subsct{g}$ is the cross-entropy loss for semantic segmentation. $w_t$ is a weighting factor defined as
\begin{equation}
w_t =
\begin{cases}
w_{\subsct{1}},& \text{$n_{t-1} < n_k$ or $t = 1$} \\
w_{\subsct{2}},& \text{$n_{t-1} \ge n_k$}
\label{w_k}
\end{cases},
\end{equation}
where $n_t$ is the number of ground truth instances at timestamp $t$. We use $w_{\subsct{1}} = 2$, $w_{\subsct{2}} = 1$ in our experiments to emphasize the changes among the adjacent frames.

MEP predicts the inter-agent events along with the visual structure as well as their uncertainty scores to enable multiple hypotheses forecasting. For example, MEP can predict the possible locations of one vehicle together with the chance of ego-vehicle colliding into it at the spot. That is, it predicts $(c, x_1, y_1, x_2, y_2, \phi)$ for each possible location, where $c$ is the prediction confidence, $(x_1, y_1, x_2,y_2)$ indicates the bounding box and $\phi$ indicates the chance of collision. If the location is very close to the ego-vehicle or in the direction the ego-vehicle is turning to, the chance of collision would be high.

\subsection{Sequential Action Sampling}
\label{sec:sas}
\begin{figure}[t]
    \centering
    \includegraphics[width=\linewidth]{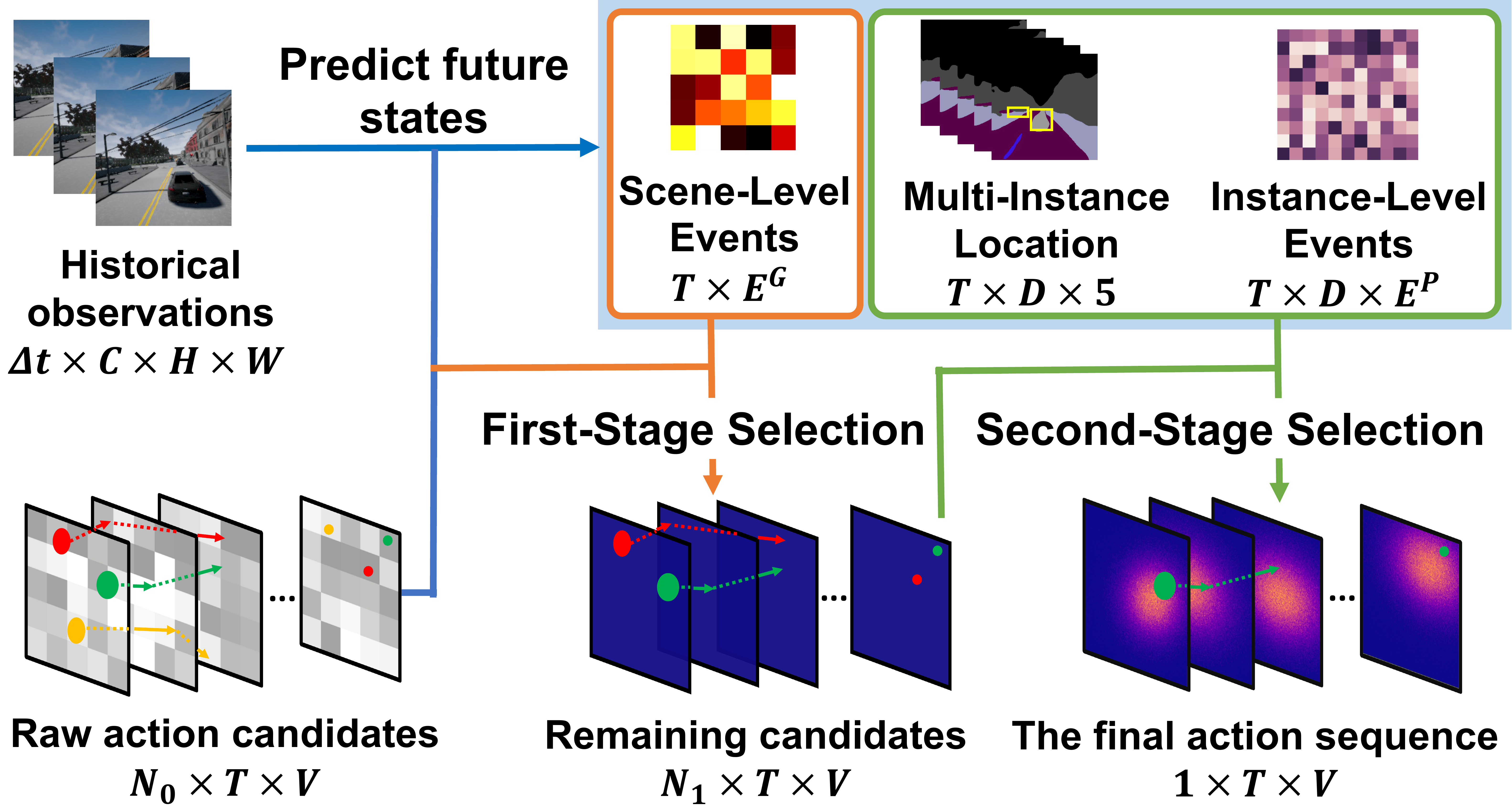}
    \caption{The procedure of SAS. From the input historical observations and sampled actions, the dynamic model predicts $E^G$ future scene-level events and $E^P$ instance-level events. From the originally sampled $N_0$ action sequences, the first stage estimates a cost referring to scene-level event chance to keep $N_1$ sequences. Then considering instance-level events, the second stage chooses the final action sequence.
    }
    \label{actionselection}
\end{figure}

\model performs sampling-based action selection based on the event prediction from the learned dynamic model. Similar to model predictive control, \model selects the action sequence from candidates that minimizes the cost of future events at each step and the ego-vehicle executes the action on the first step in the selected sequence. We propose a two-stage sequential action sampling (SAS) strategy to select the desired action sequence shown in Fig.~\ref{actionselection}. The final action is determined conditioned on the instance-level event predictions in a pruned action space by the first stage that relies on scene-level event predictions. 

\minisection{Guidance network.} We first discretize the action space into multiple bins and then uniformly sample from the selected bins by a guidance network to avoid sampling action directly from continuous space. 
At each step $t$, $N_0$ action sequence candidates with a horizon of $T$ are sampled,
and the dimension of the action space is $V$. We set $V=2$ for steering and throttle in our experiments, and both values are float between -1 and 1. For steering, -1 means turning left-most, and 1 means turning right-most. For throttle, -1 and 1 means to slow down or speed up in the full capacity.

We adopt a guidance network to output the probability of the discrete bins based on historical observations. It is trained through imitating good episodes from the replay buffer. With the imitation, the model can learn from the current trajectory and the historical experiences to improve the sample efficiency.

\minisection{Future events definitions.} With the historical observations and sampled actions, the dynamic model predicts the future states. For each action sequence, the sequential action sampling (SAS) strategy uses the predictions of both scene-level events $G$ and instance-level events $P$. In the CARLA environment, we use four scene-level events off-road, off-lane, collision and speed. We thus denote the prediction of $G$ as 
$G^{*}=(p_{\text{off-road}}, p_{\text{off-lane}}, p_{\text{collision}}, p_{\text{speed}})$. We only consider one instance-level event, inter-agent collision, at $D$ possible vehicle locations with confidence score above a threshold. The event is represented by $(c,x_1,y_1,x_2,y_2,\phi)$, where $c$ is the prediction confidence, $(x_1,y_1,x_2,y_2)$ indicates the bounding box and $\phi$ indicates the chance of collision. 

\minisection{Two-stage action sampling.} With the predicted future states, SAS selects actions to drive safely and fast. The first stage focuses on scene-level events with a cost function defined as
\vspace{-2mm}
\begin{equation}
    \cfunc_{\subsct{scene}}(\state_t, \action_{t:t+T}) = \sum^{T+1}_{i=1} \alpha_i \cfunc_{\subsct{g}}(G^{*}_{t+i}),
    \label{cost_C1}
\end{equation}
where $\alpha_i = max(\frac{1}{2^i}, \frac{1}{8})$ is the weighting factor to emphasize more on the near future prediction while not neglecting further steps. We consider three types of accidents (i.e., off-road, off-lane and collision), so the cost function $C_\subsct{g}$ is
\newcommand{\cpfunc}{\cfunc_\subsct{p}}
\vspace{-2mm}
\begin{equation}
    \cfunc_\subsct{g}(G^*) = \cpfunc(p_\subsct{off-road}) + \cpfunc(p_\subsct{collision}) + 0.2 \cpfunc(p_\subsct{off-lane}),
\end{equation}
\vspace{-2mm}
\begin{equation}
    \cpfunc(x) = - p_{\subsct{speed}} \cdot (1 - x) +  v_\subsct{m} \cdot x,
\end{equation}
where $v_\subsct{m}$ is the allowed maximum speed. This cost function encourages higher speed and lower accident chance. We initially sample $N_0$ action sequences.
Only the top $N_1$ action sequences with the minimal cost are considered in the second stage. We set $N_0=20$ and $N_1=5$ in experiments. 

The second stage focuses on instance-level events to avoid collision between the ego-vehicle and other vehicles. The cost function is then defined as 

\vspace{-2mm}
\begin{equation}
    \cfunc_\subsct{instance}(\state_t, \action_{t:t+T}) = \sum^{T+1}_{i=1} \alpha_i \sum^{D}_{d=1} \phi_{t+i,d} \cdot c_{t+i,d},
    \label{ias_cost}
\end{equation}
where $\alpha_i$ is the same weighting factor as in Eq. \ref{cost_C1}. 

\section{Experiments}

\begin{figure*}[t]
    \centering
    \includegraphics[width=0.95\linewidth]{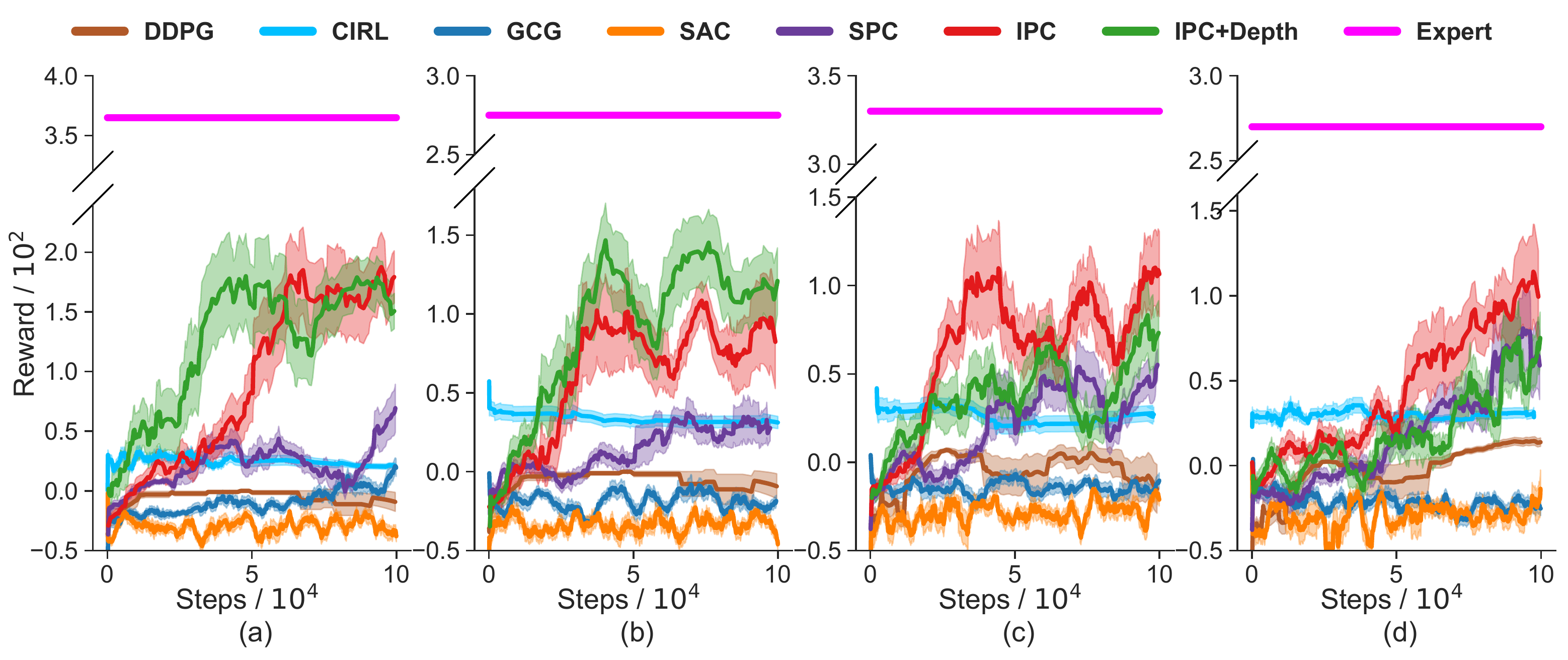}
    \caption{The reward curves of methods tested under different environment settings in CARLA. (a): \texttt{town01} w/ 32 vehicles in; (b): \texttt{town01} w/ 64 vehicles in; (c): \texttt{town02} map with 32 vehicles in; (d): \texttt{town02} w/ 64 vehicles in. In all cases, our method outperforms all other methods. The depth information helps marginally in (a) and (b).\vspace{-2mm}}
    \label{fig:benchmark}
\end{figure*}

We mainly conduct experiments in CARLA simulator. The results show \model outperforms a range of baselines (\ref{sec:comp}). We also conduct ablation studies on the novel components, MEP and SAS, and find that they can both improve performance (\ref{sec:inst}). We compare the scene-level event prediction quantitatively with the closest baseline, SPC~\cite{spc} (\ref{sec:comp_spc}). Besides, we find \model is flexible to incorporate expert demonstrations to boost model performance (\ref{sec:imit}). Finally, we show potentials to use \model in more realistic environments (\ref{sec:gta}).

\subsection{Experimental Setup}
\label{sec:exp_setup}
\minisection{Simulation environment.} We adopt the multi-agent driving environments in CARLA~\cite{Carla}, which provide multiple built-in maps with diverse scene structures and road configurations. The ego-vehicles uses a monocular RGB camera to get visual observations. 
The frame rate of the simulator is 10 frames per second.
CARLA supports extracting semantic segmentation and vehicle locations for supervision. We also evaluate our method on GTA V, which has more realistic rendering but provides no visual ground truth, so we train a Mask R-CNN~\cite{he2017mask} model on a collected dataset~\cite{gta_dataset} to provide pseudo-supervision.

\minisection{Model configurations.} We use  
Deep Layer Aggregation~\cite{DLA} to extract visual features and a single layer ConvLSTM network~\cite{convlstm} to synthesize features on future steps. MEP uses a RetinaNet~\cite{focalloss}-style head to predict instance locations and an up-sampling layer for semantic segmentation. 

\minisection{Training.} We use Adam~\cite{adam} for training with an initial learning rate $5\times10^{-4}$. The exploration rate decreases linearly from $1$ to $0$ until $100$k steps. \model forecasts on future 10 frames as in SPC~\cite{spc}. The history buffer size is $20$k, and the batch size is $16$. One episode terminates when the ego-vehicle (1) collides for $20$ times or (2) gets off-road or stuck for $30$ steps continuously, or (3) has run $1000$ steps.

\minisection{Evaluation.}
We adopt the reward function in SPC~\cite{spc} as 
\vspace{-1mm}
\begin{equation}
    \func{R}(\state_t) = \frac{\text{speed}}{15} - ( \mathds{1}_{\text{off-road}} + 2 \times \mathds{1}_{\text{collision}} + 0.2 \times \mathds{1}_{\text{offline}} ),
\end{equation}
where $\mathds{1}_{\text{off-road}}, \mathds{1}_{\text{collision}}$ and $\mathds{1}_{\text{off-lane}}$ indicate (binary) if off-road, collision or off-lane happens on that step.

\begin{table*}[t]
\begin{center}
\caption{The prediction accuracy of scene-level events by SPC~\cite{spc} and our method. The evaluation is conducted with 32 vehicles spawned in the map. Compared with SPC, our method shows more accurate prediction on all scene-level events.\label{tab:prediction_acc}}
\adjustbox{width=.8\linewidth}
{
\begin{tabular}{ c | c | ccccc ccccc } 
\toprule 
\multicolumn{2}{c|}{Horizon} & 1 & 2 & 3 & 4 & 5 & 6 & 7 & 8 & 9 & 10 \\
\midrule
\multirow{3}*{collision} & SPC & 96.12 & 96.05 & 95.74 & 95.58 & 95.41 & 95.29 & 95.03 & 94.80 & 94.67 & 94.37 \\
~ & \model & \textbf{97.34} & \textbf{97.28} & \textbf{97.17} & \textbf{97.07} & \textbf{96.86} & \textbf{96.71} & \textbf{96.52} & \textbf{96.31} & \textbf{96.07} & \textbf{95.81} \\
\midrule
\multirow{3}*{offroad} & SPC & 96.87 & 96.74 & 96.56 & 96.30 & 96.07 & 95.82 & 95.53 & 95.24 & 94.92 & 94.61 \\
~ & \model & \textbf{97.47} & \textbf{97.46} & \textbf{97.29} & \textbf{97.12} & \textbf{96.91} & \textbf{96.76} & \textbf{96.54} & \textbf{96.34} & \textbf{96.08} & \textbf{95.72} \\
\midrule
\multirow{3}*{offlane} & SPC & 94.79 & 94.89 & 94.91 & 94.63 & 94.42 & 94.05 & 93.33 & 92.82 & 92.32 & 91.51 \\
~ & \model & \textbf{95.19} & \textbf{95.18} & \textbf{95.13} & \textbf{94.82} & \textbf{94.54} & \textbf{94.28} & \textbf{93.94} & \textbf{93.53} & \textbf{93.14} & \textbf{92.61} \\
\midrule
\multirow{3}*{coll w/ veh} & SPC & 95.10 & 94.82 & 94.66 & 94.37 & 94.18 & 94.09 & 93.93 & 93.82 & 93.75 & 93.62 \\ 
~ & \model & \textbf{96.02} & \textbf{96.00} & \textbf{95.97} & \textbf{95.70} & \textbf{95.59} & \textbf{95.37} & \textbf{95.26} & \textbf{95.19} & \textbf{95.02} & \textbf{94.96} \\ 
\bottomrule
\end{tabular}}
\end{center}
\end{table*}

\begin{table*}[t]
\begin{center}
\caption{Semantic segmentation prediction accuracy in single-agent (SA) and multi-agent (MA) environments with 32 vehicles spawned. Though the prediction is accurate in single-agent environments, it becomes more unreliable in multi-agent environments with more uncertainties. And the semantic prediction for future vehicles is extremely inaccurate. It shows semantic prediction is not enough to drive in multi-agent environments.\label{tab:semantic_acc}}
\adjustbox{width=.8\linewidth}
{
\begin{tabular}{ c|c | ccccc ccccc } 
\toprule 
\multicolumn{2}{c|}{Horizon} & 1 & 2 & 3 & 4 & 5 & 6 & 7 & 8 & 9 & 10 \\
\midrule
\multirow{4}*{SA} & pixel acc. & 96.50 & 95.45 & 95.39 & 95.25 & 95.19 & 95.14 & 95.06 & 95.02 & 94.91 & 94.70 \\
~ & mean acc. & 95.95 & 95.75 & 95.60 & 95.51 & 95.43 & 95.27 & 95.15 & 94.97 & 94.76 & 94.43 \\
~ & mean IU & 93.45 & 92.76 & 92.50 & 92.32 & 92.21 & 92.07 & 91.94 & 91.77 & 91.58 & 91.39 \\
~ & f.w. IU & 94.38 & 94.11 & 93.88 & 93.67 & 93.49 & 93.30 & 93.17 & 93.02 & 92.86 & 92.65 \\
\midrule
\multirow{4}*{MA} & pixel acc. & 94.52 & 93.27 & 92.54 & 91.52 & 91.30 & 90.72 & 90.38 & 90.07 & 89.90 & 89.53 \\
~ & vehicle acc. & 48.67 & 42.90 & 42.62 & 41.83 & 41.04 & 39.36 & 39.01 & 38.76 & 38.50 & 38.27\\
~ & mean acc. & 80.21 & 73.20 & 71.12 & 70.41 & 68.50 & 67.36 & 67.07 & 66.83 & 66,07 & 65.50\\
~ & mean IU & 71.96 & 66.42 & 64.60 & 62.97 & 62.01 & 61.60 & 60.99 & 60.32 & 59.54 & 59.02 \\ 
~ & f.w. IU & 90.56 & 88.08 & 86.81 & 84.92 & 84.60 & 84.01 & 83.76 & 83.21 & 82.77 & 81.74 \\ 
\bottomrule
\end{tabular}}
\vspace{-4mm}
\end{center}
\end{table*}

\begin{table}
    \centering
    \caption{The statistical results of driving on Twon01. Frequency is calculated on every 100 steps.}
    \adjustbox{width=.9\linewidth}{
    \begin{tabular}{c|cccc}
        \toprule
        & avg speed & collision & coll w/ veh  & offroad\\
        \midrule
        IPC-32 & 10.22 & 1.23 & 1.07 & 1.12 \\
        SPC-32 & 8.79 & 3.02 & 2.79 & 2.23  \\
        \midrule
        IPC-64 & 9.32 & 2.53 & 2.23&  1.45 \\
        SPC-64 & 8.03 & 4.12 & 3.98 & 2.38 \\
        \bottomrule
    \end{tabular}}
    \label{tab:raw_data}
\end{table}

\begin{figure*}
    \centering
    \subfigure[]{
    \begin{minipage}[t]{0.45\linewidth}
    \centering
    \includegraphics[width=\linewidth]{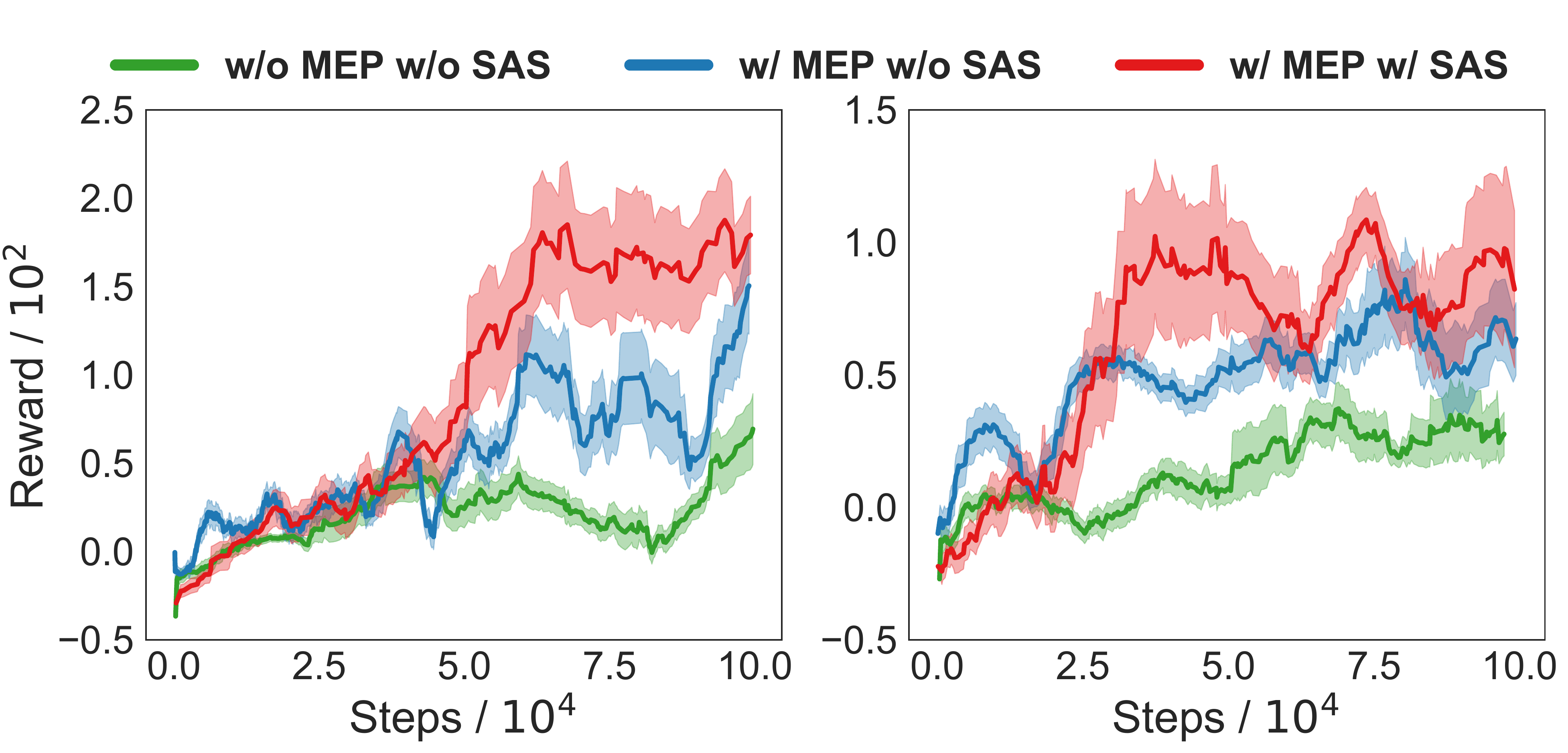}
    \end{minipage}%
    }%
    \subfigure[]{
    \begin{minipage}[t]{0.45\linewidth}
    \centering
    \includegraphics[width=\linewidth]{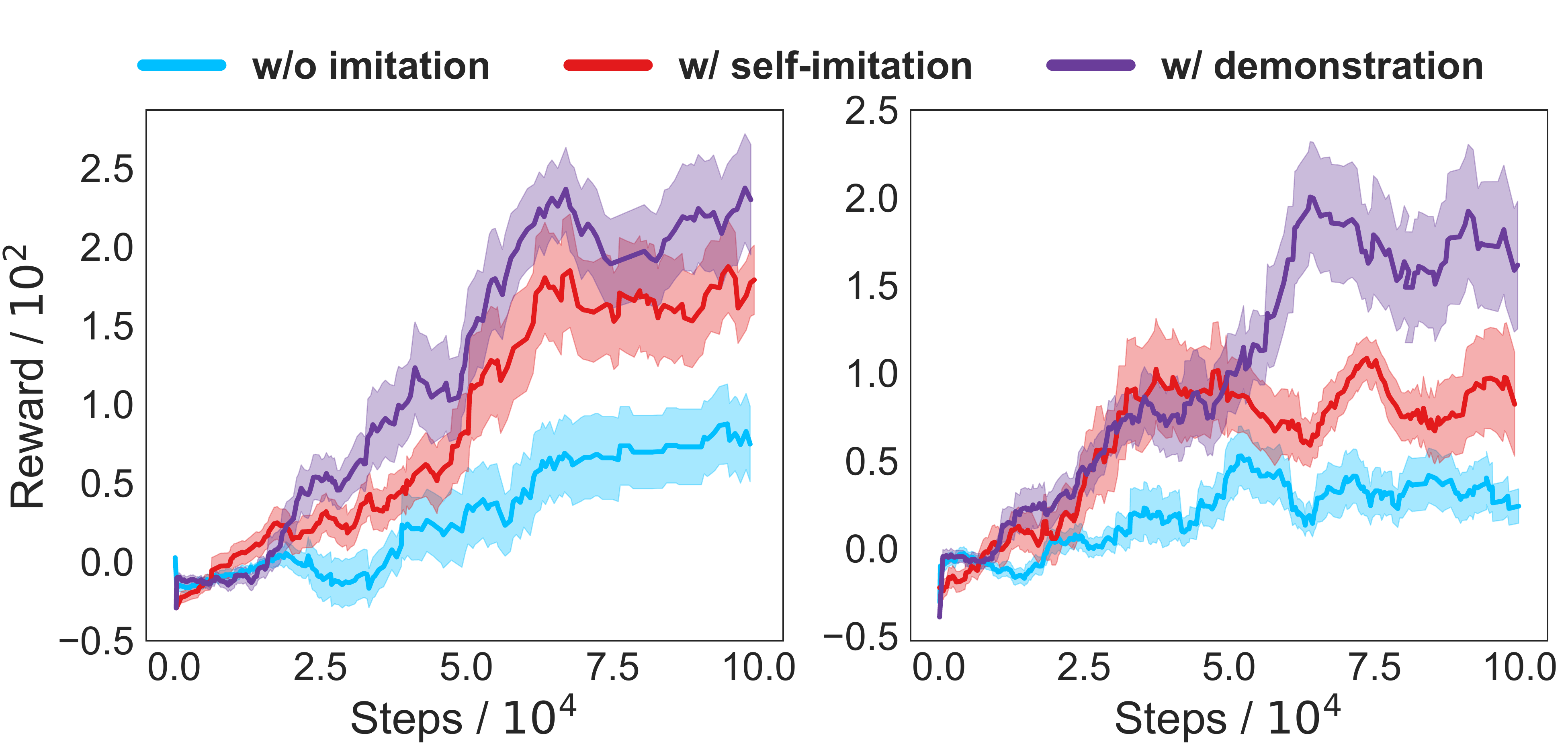}
    \end{minipage}%
    }%
    \caption{(a). Ablation study of MEP and SAS. It is tested on \texttt{town01} map with 32 vehicles (left) or 64 vehicles (right) spawned. The results show that MEP improves the data efficiency a lot and SAS makes better action decisions. (b). How imitation boosts the model performance with 32 vehicles (left) or 64 vehicles (right) spawned. The results show that demonstration and historical experience both boost the model performance but demonstration helps more than the pattern from good experience.
    }
    \label{fig:ablation_study}
\end{figure*}

\subsection{Results in CARLA}
\label{sec:comp}
We compare \model with both model-free (DDPG~\cite{DDPG} and SAC~\cite{SAC}) and model-based methods (GCG~\cite{GCG} and SPC~\cite{spc}). DDPG and SAC output action signals directly from visual observations. SPC and GCG do predictive control similar to \model, but SPC uses visual signal while GCG does not. We also use a depth map as an extra supervision signal to study if it can boost the learning rewards. Our comparison also includes CIRL~\cite{CIRL}, which uses imitation learning to pretrain model weights and finetunes with DDPG.
An expert autopilot agent is deployed to give a reward reference for accident-free driving. 
As shown in Fig.~\ref{fig:benchmark},  \model obtains higher rewards than 
baselines on two maps.
More agents in the environments make it harder to drive, so all the methods' rewards deteriorate even for the expert agent. Despite the challenges, our model still performs well in 64-agent environments. We find \model is  more sample efficient.

\subsection{\model v.s. SPC}
\label{sec:comp_spc}

We examine the differences between \model and SPC on CARLA. 
We compare their scene-level event prediction accuracy in Table~\ref{tab:prediction_acc}. We find that \model outperforms SPC on all four scene-level event predictions within a horizon of 10 steps. We hypothesize that the scene structure prediction using semantic segmentation in multi-agent environments maybe not reliable anymore. We, therefore, test the semantic segmentation prediction accuracy of \model in single-agent and multi-agent environments in 
Table~\ref{tab:semantic_acc}.  It shows that the accuracy drops substantially in multi-agent environments, indicating the necessity of using instance-level cues. We also compare raw driving metrics for SPC and \model in Table~\ref{tab:raw_data} which agree that \model makes the ego-vehicle drive faster and more safely in multi-agent environments. Accidents cause a risk for the ego-vehicle to get into situations it would not have predicted. So by better avoiding accidents, \model forecasts future states more accurately.


\subsection{Ablation study of MEP and SAS}
\label{sec:inst}
MEP estimates instance-level events related to other agents, providing cues for more reliable control.
SAS selects action with better tolerance of forecasting noise.
We conduct an ablation study for the two modules in 
Fig.~\ref{fig:ablation_study}(a). 
As the figure shows,
the reward increase is slower when MEP is not used. This indicates MEP can improve the sample efficiency. Moreover, without SAS, we merge two stages into one. It shows that the performance is lower than SAS. It indicates that SAS helps to select more reliable actions.

\begin{figure}
    \centering
    \includegraphics[width=0.95\linewidth]{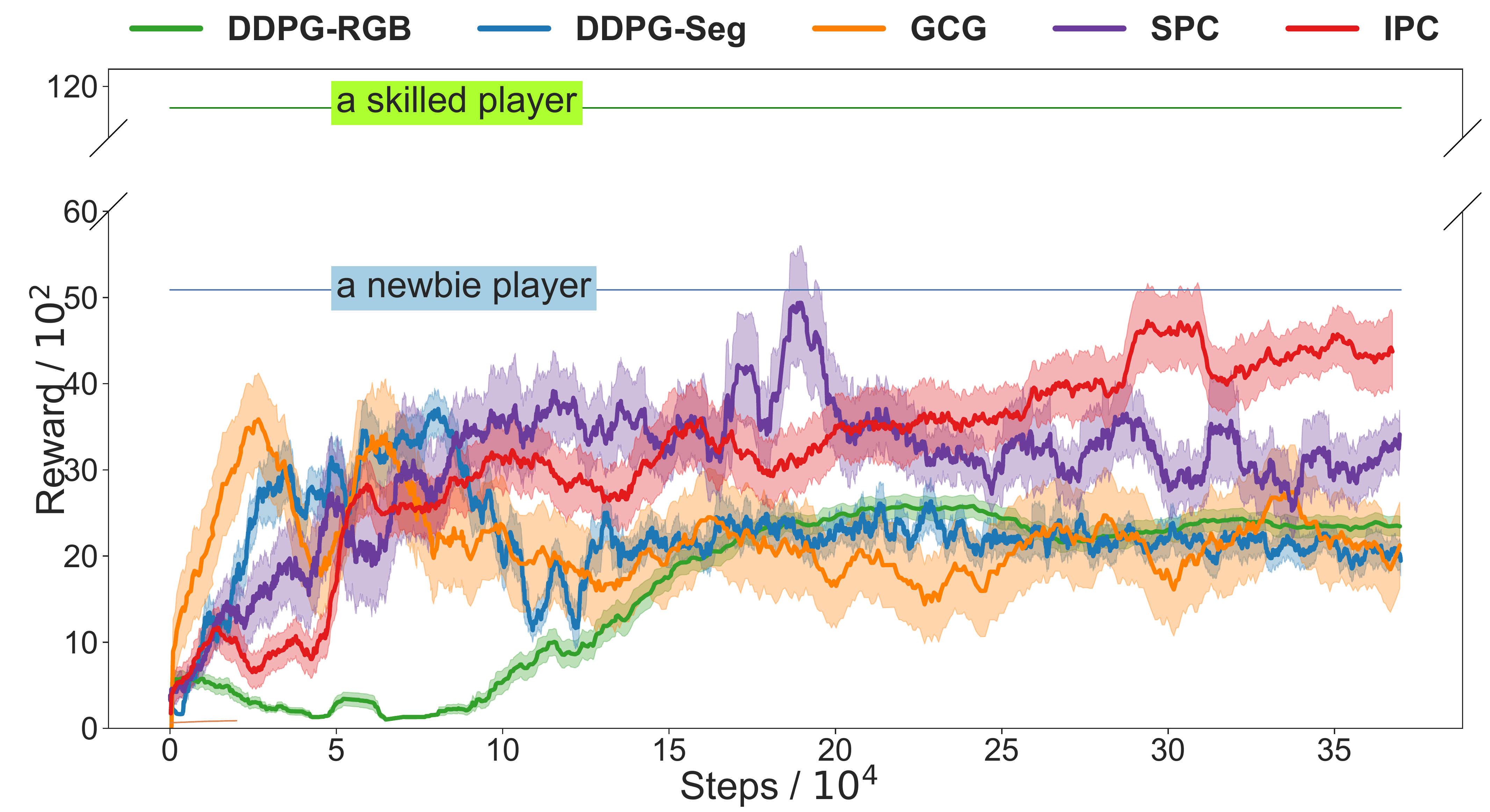}
    \caption{Results on GTA V. \model outperforms other methods.
    }
    \label{fig:gta}
\end{figure}

\subsection{Boost Performance by Imitation}
\label{sec:imit}

So far, we have shown that \model achieves competitive results in multi-agent environments without any expert demonstration. Yet, \model is flexible to incorporate expert
demonstrations. 
Imitation can be used in training the guidance network to sample actions.
We utilize expert demonstrations from the autopilot agent in CARLA for imitation. We also have self-imitation learning to imitate actions taken in historical good episodes, in which reward is at the top 10\% and higher than 200. Results are shown in Fig.~\ref{fig:ablation_study}(b). It shows the expert demonstration improves
sample efficiency not substantially compared to the self-imitation
when there are 32 vehicles. But the model benefits much more from demonstrations when there are 64 other vehicles.  With the ability to learn from imitations, we can collect demonstrations from either simulators or the real world and boost model performance offline. Joint optimizing the visual prediction and action decision making forms a scalable path to learn more robust action patterns among different environments.

\subsection{Evaluation on more realistic benchmarks}
\label{sec:gta}
Though CARLA is designed for evaluating driving algorithms and systems, its image rendering realism still has space for improvement compared to the latest rendering technology. Therefore, we also evaluate our method on GTA V, a video game with more realistic rendering. We run DDPG, GCG, SPC, and \model in GTA V for the comparison. As a reference of human performance, we also show the driving rewards of a skilled player, who has rich experience in playing GTA V, and a newbie player, who has not played GTA V. The reference rewards are their average performance with 10 trials. As shown in Fig.~\ref{fig:gta}, \model outperforms all other baselines. Moreover, both \model and SPC can have a close performance with the newbie player, but \model is more stable.

\begin{figure}
    \centering
        \begin{tabular}{cc}
            \includegraphics[width=4cm]{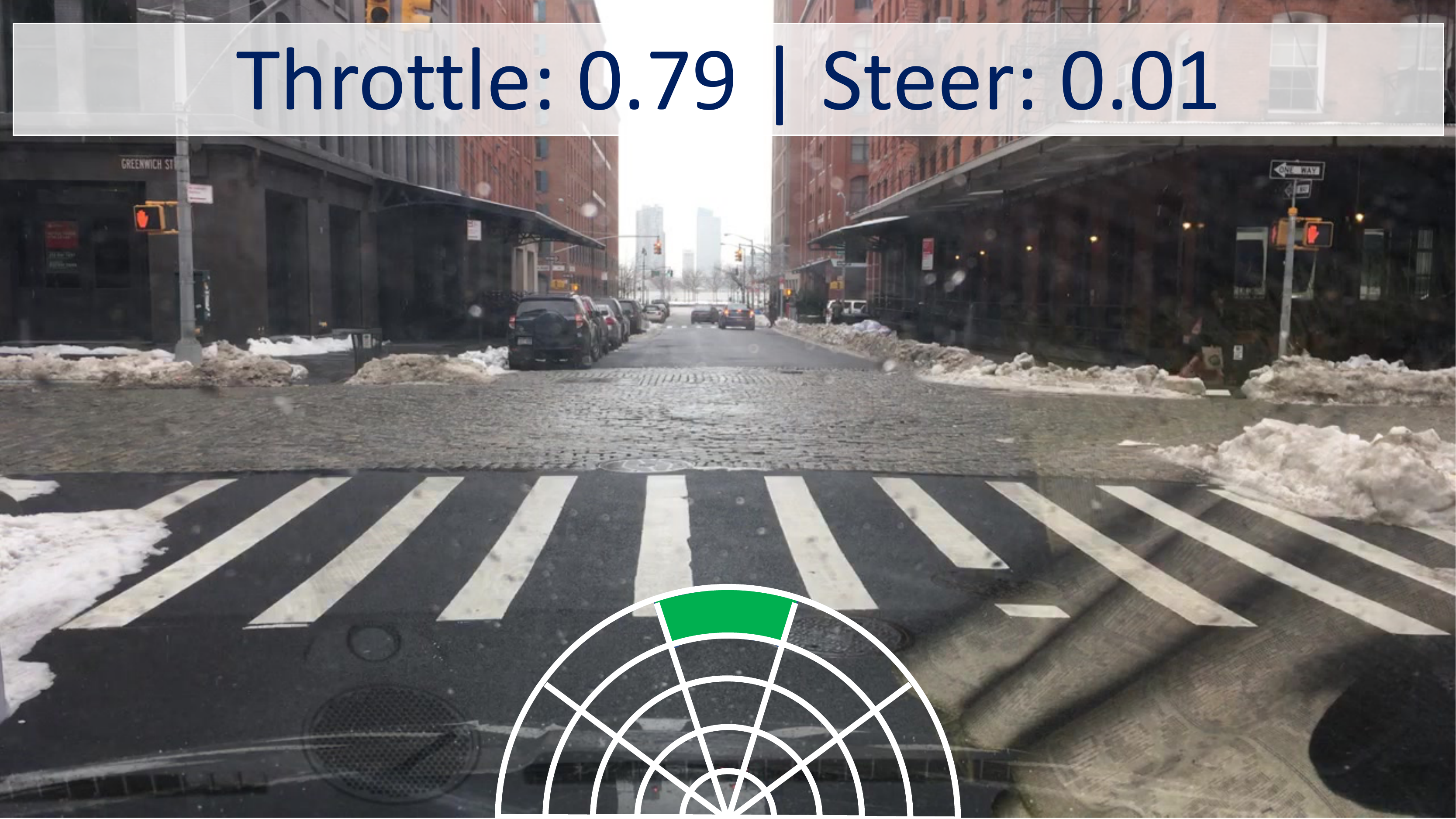}& \includegraphics[width=4cm]{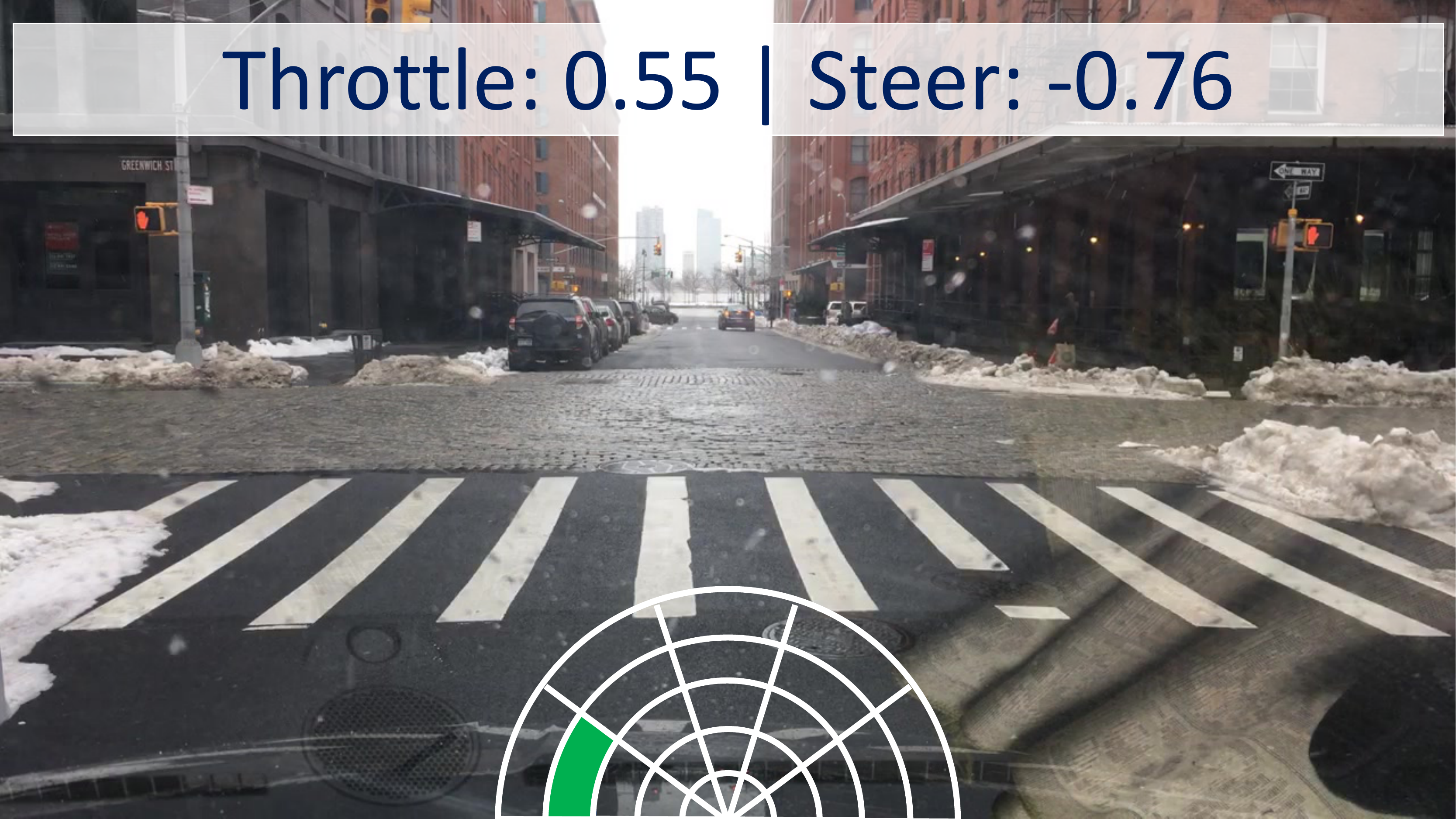}
            \\
            (a) & (b)\\
            \includegraphics[width=4cm]{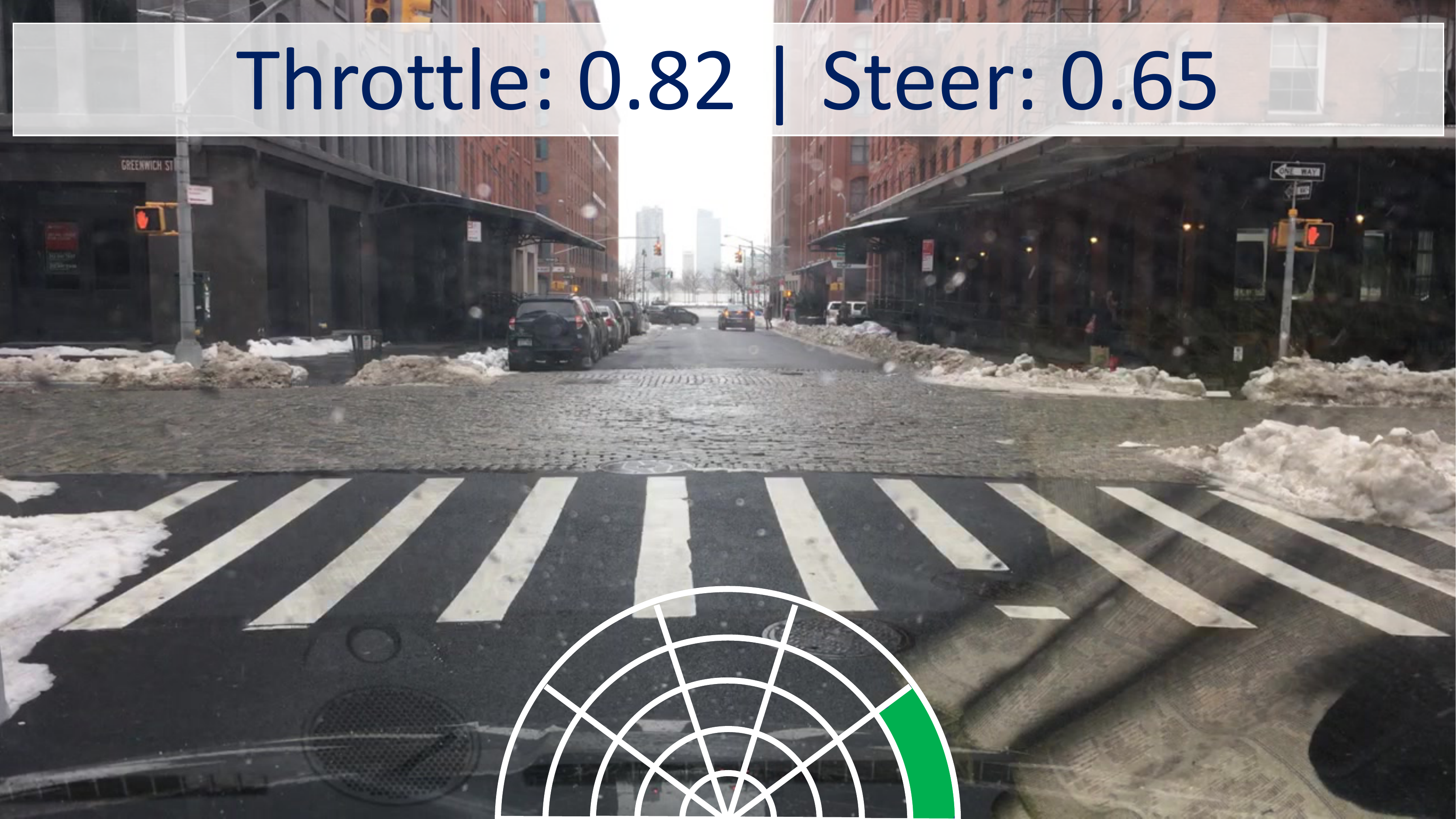}& \includegraphics[width=4cm]{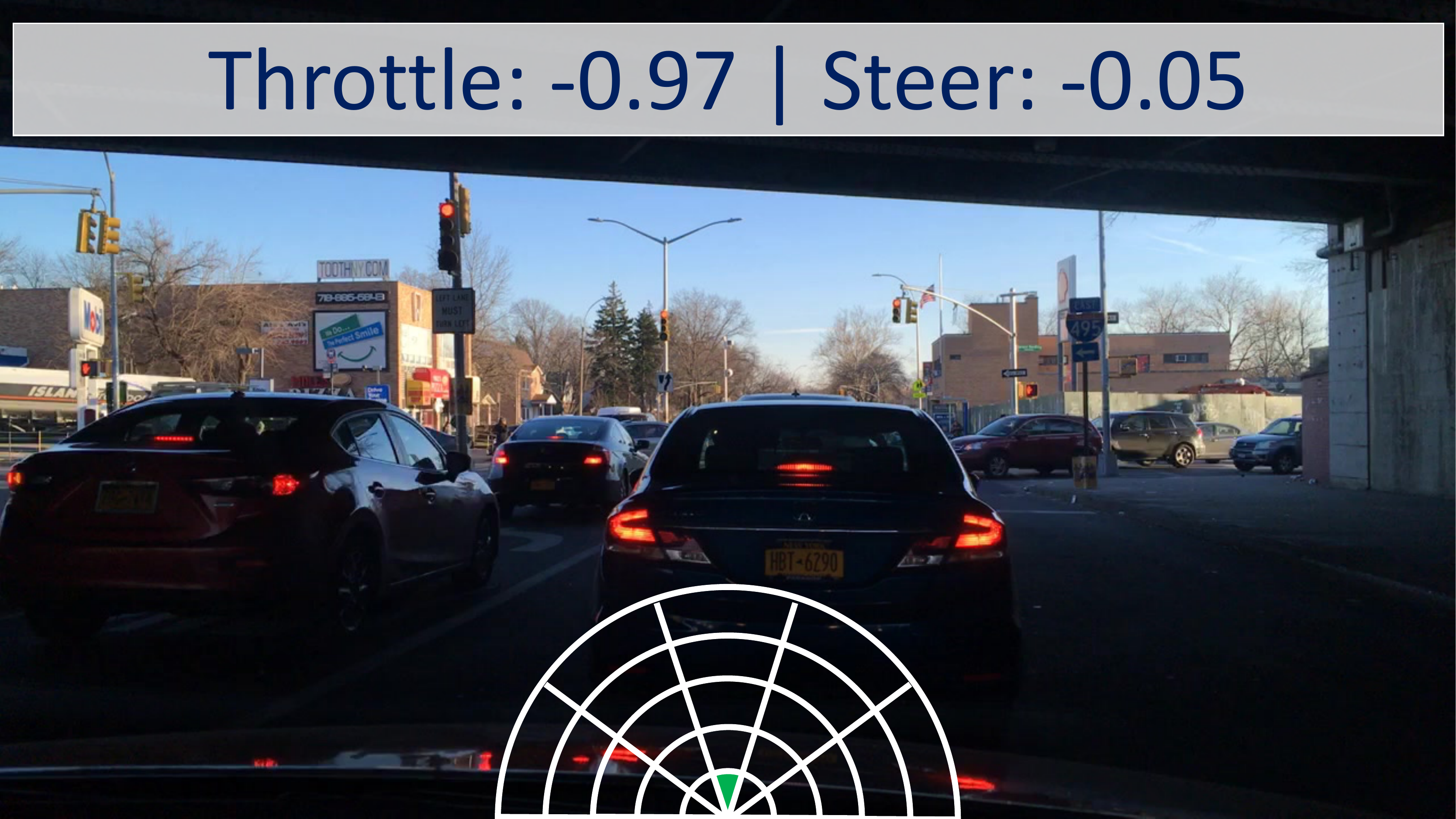}
            \\
            (c) & (d)\\
            \includegraphics[width=4cm]{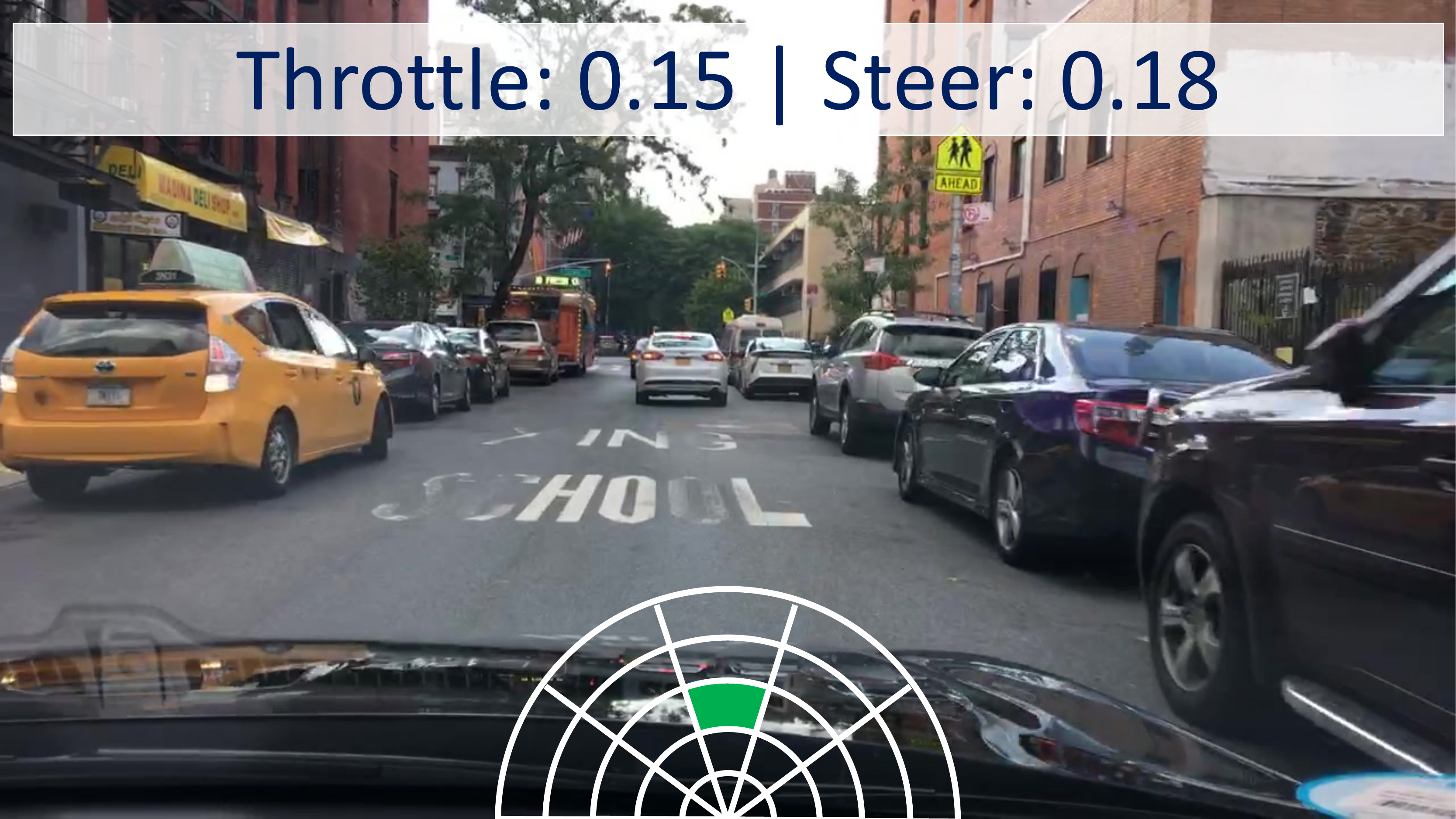}& \includegraphics[width=4cm]{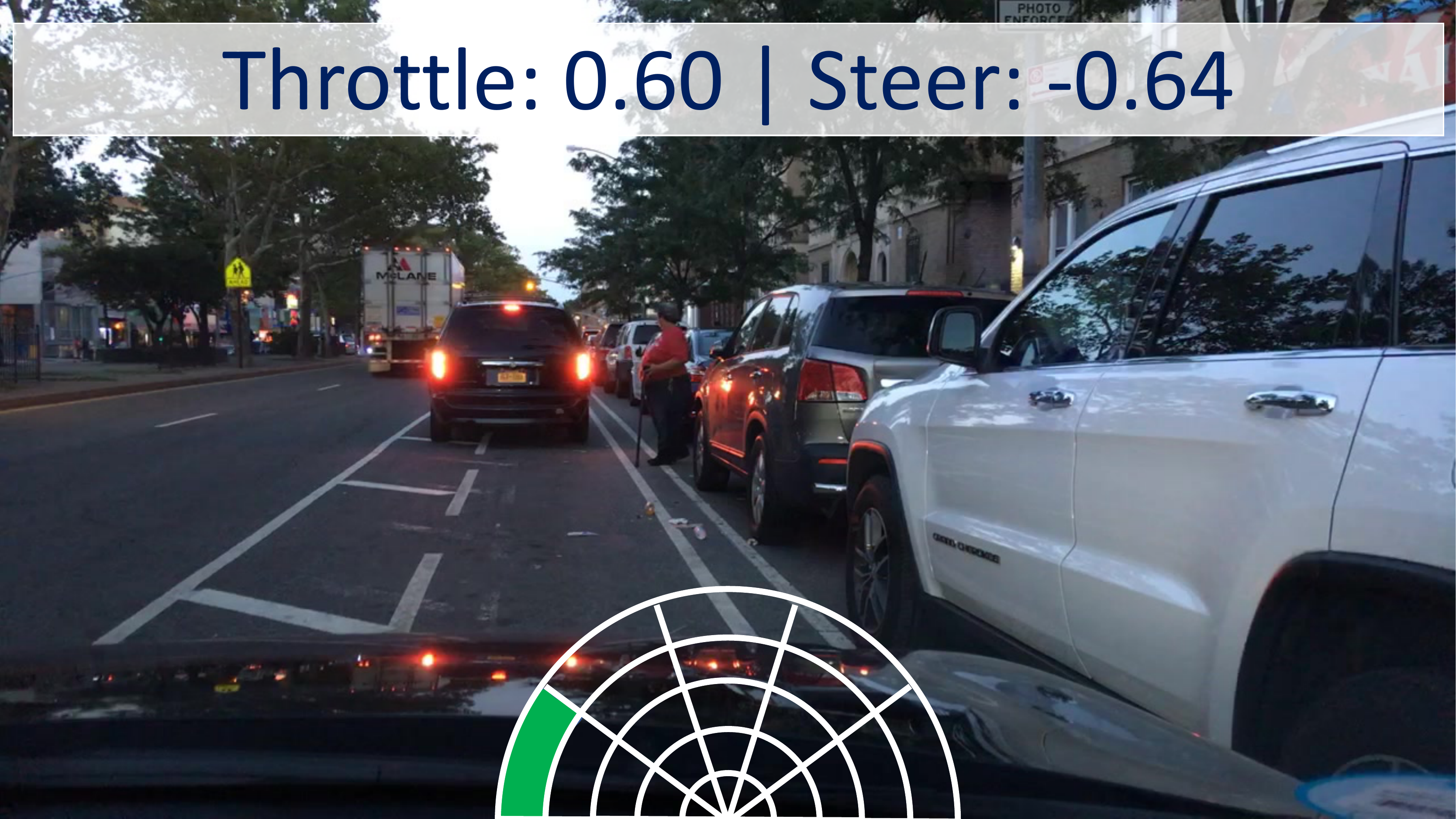}
            \\
            (e) & (f)\\
        \end{tabular}
    \caption{The policy from simulator shows reasonable action even in real-world scenes. In (a)-(c), it is confident to drive to three directions in a crossing. In (d), it slows down to wait for the vehicle ahead. In (e), it follows straight on the road. In (f), it drives towards left to avoid the vehicle on the right. \vspace{-8mm}
    }
\label{fig:bdd}
\end{figure}

It is challenging to adapt the model learned from the simulator to real-world datasets. But we still try to directly deploy the model trained on CARLA to the real-world driving video dataset BDD100K~\cite{yu2020bdd100k}. It generates reasonable actions in some scenes, as shown in Fig.~\ref{fig:bdd}, but can fail a lot. Our study provides some basis for a better sim-to-real policy transfer in the future.

\section{Conclusion}
We presented an instance-aware predictive control (IPC) approach for driving policy learning in multi-agent environments. 
\model includes a novel multi-instance event prediction (MEP) module to forecasts future events at both scene and instance levels. MEP predicts the scene structure changes through semantic segmentation and the possible locations of other vehicles.
To efficiently select actions conditioned on the multi-level event prediction, we introduced a two-stage sequential action sampling strategy to improve control from noisy forecasting. Experiments on realistic driving simulation environments showed that our method built new state-of-the-art under different multi-agent settings with improved data efficiency and policy reliability.

%
%

\bibliographystyle{splncs04}
\bibliography{bib}


\end{document}